\documentclass[twocolumn]{article}
\usepackage{authblk}
\usepackage[bookmarks, pdftex, colorlinks=true, pagebackref=true, backref=page]{hyperref}
\usepackage[square,numbers]{natbib}
\newcommand{\state}{}

\usepackage[utf8]{inputenc} %
\usepackage[T1]{fontenc} %
\usepackage{type1cm} %
\usepackage{amsmath} %
\usepackage{balance} %
\usepackage{booktabs} %
\usepackage{bold-extra} %
\usepackage[font={bf}, tableposition=top]{caption} %
\usepackage{graphicx} %
\usepackage{adjustbox} %
\usepackage{hyphenat} %
\usepackage{mathtools} %
\usepackage{microtype} %
\usepackage{multirow} %
\usepackage{siunitx} %
\usepackage{xfrac} %
\usepackage{xspace} %
\usepackage{cleveref} %
\PassOptionsToPackage{hyphens}{url} %
\PassOptionsToPackage{bookmarks, pdftex, colorlinks=true, pagebackref=true, backref=page}{hyperref} %
\PassOptionsToPackage{square,numbers}{natbib} %
\usepackage[show]{chato-notes} %
\usepackage{placeins}
\usepackage[bb=dsserif]{mathalpha}
\usepackage{bm}
\usepackage[show]{chato-notes}

\newenvironment{squishlist}
{\begin{list}{$\bullet$}
 {\setlength{\itemsep}{0pt}
     \setlength{\parsep}{3pt}
     \setlength{\topsep}{3pt}
     \setlength{\partopsep}{0pt}
     \setlength{\leftmargin}{1.5em}
     \setlength{\labelwidth}{1em}
     \setlength{\labelsep}{0.5em} } }
{\end{list}}

\usepackage{xcolor}
\hypersetup{
   bookmarks, pdftex,
   colorlinks=true,
   pagebackref=true, backref=page,
   linkcolor={red!50!black},
   filecolor={green!50!black},
   citecolor={green!50!black}, 
   urlcolor={blue!80!black},
}

\graphicspath{{img/}}

\DeclareMathOperator*{\argmax}{arg\,max}
\newcommand{\bcm}{\text{BCM}\xspace}
\newcommand{\bcmlong}{{Bounded-Confidence Model}\xspace}
\newcommand{\cb}{\ensuremath{\epsilon}\xspace}
\newcommand{\da}{\text{DA}\xspace}
\newcommand{\dalong}{{Data Assimilation}\xspace}

\newcommand{\error}{\ensuremath{\mathcal{E}}\xspace}

\newcommand{\neighbors}{\ensuremath{\Gamma}\xspace}
\newcommand{\lbi}{\text{LBI}\xspace}
\newcommand{\lbilong}{{Likelihood-Based Inference}\xspace}
\newcommand{\rate}{\ensuremath{\mu}\xspace}
\renewcommand{\state}{\ensuremath{\mathbf{x}}\xspace}
\newcommand{\statei}[1]{\ensuremath{{x}_{#1}}\xspace}

\newcommand{\obs}{\ensuremath{\mathbf{y}}\xspace}
\newcommand{\obsi}[1]{\ensuremath{y_{#1}}\xspace}
\newcommand{\xest}{\ensuremath{\widehat{\state}}\xspace}
\newcommand{\xesti}[1]{\ensuremath{\widehat{\statei{#1}}}\xspace}
\newcommand{\xestf}{\ensuremath{\check{\state}}\xspace}

\newcommand{\rev}[1]{\textcolor{black}{#1}\xspace}

\title{Comparing Data Assimilation and Likelihood-Based Inference on Latent State Estimation in~Agent-Based~Models}

\author[1]{Blas Kolic}
\author[2]{Corrado Monti}
\author[3]{Gianmarco~De~Francisci~Morales}
\author[2]{Marco~Pangallo}
\affil[1]{{Universidad Carlos III},{Ronda de Toledo, 1}, {28005}, Madrid, {Spain}}
\affil[2]{{CENTAI}, {Corso Inghilterra, 3}, {10138}, Turin, {Italy}}
\affil[3]{{Intesa Sanpaolo Innovation Center}, {Corso Inghilterra, 3}, {10138}, Turin, {Italy}}

\begin{document}

\maketitle

\abstract{
In this paper, we present the first systematic comparison of \dalong (\da) and \lbilong (\lbi) in the context of \rev{an Agent-Based Model (ABM)}.
These models generate observable time series driven by evolving, partially-latent microstates.
Latent states must be estimated to align simulations with real-world data, a task traditionally addressed by \da, particularly in continuous and equation-based models used in weather forecasting.
However, the nature of ABMs poses challenges for standard \da methods.
Solving such issues requires adapting previous \da techniques or using ad hoc alternatives such as \lbi.
\da approximates the likelihood in a model-agnostic way, making it broadly applicable but potentially less precise.
In contrast, \lbi provides more accurate state estimation by directly leveraging the model's likelihood, but at the cost of requiring a hand-crafted, model-specific likelihood function, which may be complex or infeasible to derive.
We compare the two methods on the \bcmlong, a well-known opinion dynamics ABM, where agents are affected only by others holding sufficiently similar opinions.
We find that \lbi better recovers latent agent-level opinions, even under model mis-specification, leading to improved individual-level forecasts.
At the aggregate level, however, both methods perform comparably, and \da remains competitive across levels of aggregation under certain parameter settings.
Our findings suggest that \da is well-suited for aggregate predictions, while \lbi is preferable for agent-level inference.
}

\textbf{Significance statement:} Agent-based models (ABMs) are increasingly used to study real-world systems, yet their empirical calibration on latent states remains challenging and underexplored.
We provide the first systematic comparison of Data Assimilation (DA) and Likelihood-Based Inference (LBI) for latent state estimation in \rev{an ABM}. 
Our results \rev{on the Bounded-Confidence Model, a well-studied ABM,} show that while LBI offers significantly greater precision at the individual level, DA reliably captures aggregate dynamics without requiring model-specific likelihoods. 
\rev{This result suggests that} when agent-level accuracy is essential, LBI is preferable, but for \rev{complex models} where the true likelihood is difficult or impossible to specify, DA provides a robust alternative. 

\section{Introduction}
\label{sec:intro}

Agent-Based Models (ABMs) have become indispensable tools for studying complex systems across various disciplines, including economics, epidemiology, ecology, and sociology \citep{wilensky2015introduction,railsback2019agent,pangallo2023unequal,starnini2025opinion}.
By explicitly representing individual agents, each with distinct behaviors, interactions, and adaptive rules, ABMs capture how macro-scale patterns emerge from micro-level heterogeneity.
This granularity enables researchers to explore counterfactual scenarios, test policy interventions, and uncover mechanisms driving phenomena such as market crashes, disease spreading, or cultural shifts.
Unlike aggregate models, ABMs preserve the interpretability of individual decisions while accommodating nonlinearity and path dependence.
Their flexibility makes them particularly valuable in the social sciences, where human behavior often defies simplistic averaging assumptions.

A paradigmatic example is opinion dynamics, where ABMs such as the \bcmlong (\bcm) simulate how agents influence one another's views~\citep{lorenz2007continuous}.
In the \bcm, agents iteratively adjust their opinions only when interacting with others whose beliefs lie within a fixed confidence bound~\citep{deffuant2000mixing}.
This simple rule generates rich macro-level outcomes, such as polarization, fragmentation, or consensus, that can be understood in terms of single agent-level trajectories.

Despite their simplicity, calibrating such models to real-world data poses a critical challenge: while interactions and their outcomes may be observable, the latent microstates---the evolving opinions of agents---are typically inaccessible.
Here, we interpret calibration as the broad task of aligning a simulation-based model to real-world data~\citep{brenner2007taxonomy,windrum2007empirical,lux2018empirical}, whether through estimating or tuning a few \textit{global} parameters ~\citep{grazzini2015estimation,platt2020comparison,kim2021automatic,benedetti2022black,lenti2024likelihoodbased,lenti2024variational,pangallo2024data}, or through initializing and tracking agent-level latent attributes and variables ~\citep{monti2020learning,tang2022data,kolic2022estimating,lenti2024variational}.
Traditionally, most research has focused on parameter estimation~\citep{windrum2007empirical,fagiolo2019validation,cranmer2020frontier}, largely due to a lack of appropriate methodologies for estimating latent states.
Yet, both parameter tuning and latent state estimation are paramount to the broader calibration goal.
Reconstructing the latent states is thus essential for both validating ABMs, interpreting mechanisms, and generating accurate forecasts \citep{pangallo2024data}.

Latent state inference in ABMs faces three key challenges.
First, the high dimensionality of microstates (e.g., opinions of thousands of agents) complicates inference, especially when observations are sparse or aggregate.
Second, ABMs often blend deterministic rules (e.g., the \bcm's interaction threshold) with stochastic elements, creating hybrid dynamics that resist traditional analytical methods.
Third, observation granularity varies widely: real-world data may capture single interactions (e.g., social network ties), agent-level summaries (e.g., individual survey responses), or population-level statistics (e.g., polling averages), each requiring distinct inference approaches.

To address these challenges, two methodologies have recently gained traction: \lbilong (\lbi), which optimizes parameters and states against a probabilistic model \rev{with a tractable likelihood}~\citep{monti2020learning,monti2023learning}, and \dalong (\da), \rev{which sequentially incorporates observations into model simulations to produce an optimized estimate of the system’s true state under an approximate likelihood specification \citep{kolic2022estimating}. We implement \da using the Ensemble Kalman Filter (EnKF), a Monte Carlo method that approximates the likelihood under the assumption of locally linearizable dynamics with additive Gaussian noise, yet has proven effective in high-dimensional nonlinear systems, including ABMS \citep{vetra2018state, cocucci2022inference, oswald2025agent, suchak2024coupling}.} 
A systematic comparison of these methods \rev{within a single agent-based model} is missing, leaving practitioners without guidance on their trade-offs.
In this paper, we bridge this gap by evaluating \da and \lbi for latent state estimation and forecasting in the \bcm.
We focus on three questions:
\begin{squishlist}
    \item \textbf{State Recovery}: Can \da and \lbi accurately reconstruct agent-level latent opinions?
    In a system where the state evolution equations are deterministic, such as the \bcm, this task is equivalent to estimating the initial opinions of the agents.
    \item \textbf{Forecasting Accuracy}: How does latent state recovery affect forecasting errors for observable variables (i.e., interactions)?
    This task is challenging in systems with feedback loops, where inaccuracies in initial estimates can propagate nonlinearly.
    \item \textbf{Robustness}: How do these methods perform under model mis-specification, such as noise-corrupted states or erroneous confidence bounds?
\end{squishlist}

Our results \rev{in this case study} show that \lbi outperforms \da in recovering agent-level latent states, even under mis-specification, which in turn leads to better individual-level forecasts.
However, both methods perform similarly at the aggregate level, suggesting that \da may suffice for macro-scale predictions while \lbi is preferable for micro-level inference.
This degree of reliability is striking since DA operates without model-specific likelihoods.
These findings \rev{are a first step towards building a concrete,} practical guidance for ABM calibration and highlight trade-offs between methodological complexity and accuracy.

\section{Results}
\label{sec:results}

We evaluate the performance of both \dalong (\da) and \lbilong (\lbi) on a deterministic, synchronous variant of the \bcmlong (\bcm) of opinion dynamics~\citep{deffuant2000mixing} running on a fully-connected network.
Agents iteratively converge in their opinions based on interactions with neighbors whose opinions lie within a fixed, known confidence bound \cb, following the deterministic update rules of \citet{deffuant2000mixing}.
The steady state may result in either consensus, polarization, or fragmentation, depending on $\cb$ and the initial conditions.

To evaluate latent state recovery, we infer agents' opinions using observed data from the first $25\%$ of a simulated trajectory (\num{250} out of \num{1000} time steps).
Reconstruction accuracy is quantified by comparing estimated opinions to ground truth at the final observed timestep ($t=250$).
The \bcm's deterministic dynamics reduce state recovery to estimating the initial conditions $\state(0)$.
Forecasting performance is evaluated by initializing the BCM with the reconstructed states at $t=250$, simulating forward for $t>250$, and considering the observable variable, i.e., the interactions.
Forecast errors are measured at three aggregation levels: edge-level accuracy in predicting pairwise interactions, agent-level deviation in the number of interactions with its neighborhood, and graph-level discrepancies in macro-scale aggregate metrics, which indicate polarization or consensus.

We further assess the robustness of the methods by introducing two forms of mis-specification: ($i$) noise-corrupted states, where stochasticity is artificially injected into the latent opinion update, and ($ii$) incorrect confidence bounds, where the assumed value of \cb during inference is incorrect.

We evaluate the ability of \da (blue in the figures) and \lbi (red) to reconstruct latent opinion trajectories in the deterministic \bcmlong (\bcm), using only agent-level interaction events as observations.
We assess their performance in reconstructing the agent opinions and in forecasting future agent interactions using the reconstructed state.

\subsection{Latent State Inference}
\label{sec:inference}

\begin{figure*}[h!]
    \centering
    \includegraphics[width=1\textwidth]{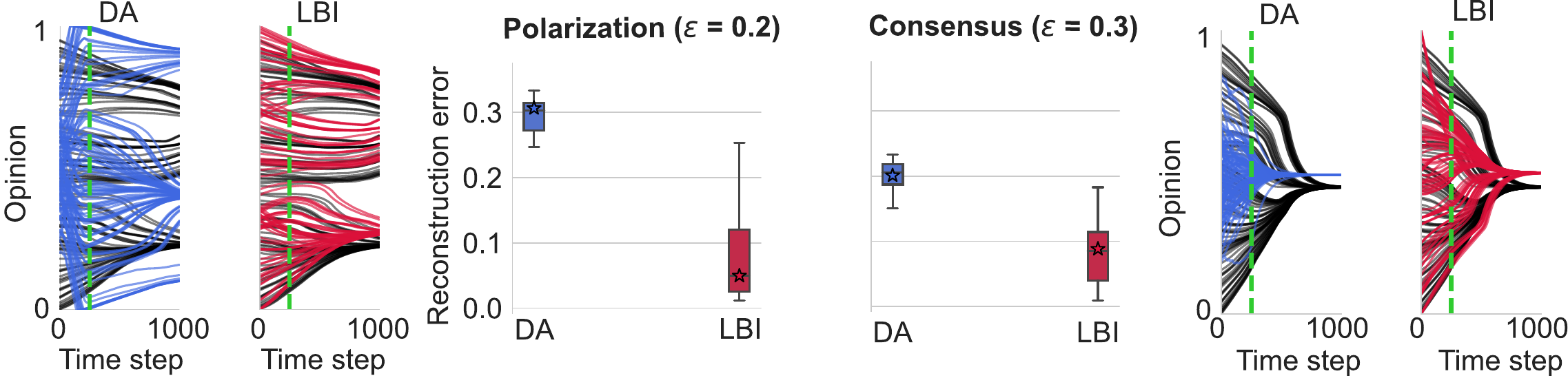}
    \caption{Estimation results for \lbi and \da for two levels of $\epsilon$. 
    The box plots in the center represent the reconstruction error on the y-axis at the end of training ($T=250$) for \da and \lbi (x-axis).
    Beside each box, we depict one of the corresponding traces, with the true (in black) and estimated (blue for \da, red for \lbi) positions in one single estimation experiment (whose error is represented with a star in the bar plot, for reference). In each trace plot, the x-axis represents time, and the y-axis represents the opinion of each agent.
    }
    \label{fig:reconstruction_clean}
\end{figure*}

In \Cref{fig:reconstruction_clean}, we compare reconstructions of noise-free BCM trajectories under two regimes: \emph{polarization} ($\cb = 0.2$) and \emph{consensus} ($\cb = 0.3$). 
Across both regimes, \lbi consistently recovers latent opinions more accurately than \da. 
This can be seen qualitatively in the closer alignment of reconstructed and true trajectories, and quantitatively in the reconstruction errors. 
For polarization, the error is substantially lower under \lbi (mean $=0.09$, Interquartile Range (IQR) $=0.04\text{--}0.13$) compared to \da (mean $=0.28$, IQR $=0.26\text{--}0.31$). 
Similarly, for consensus, \lbi achieves a mean error of $0.09$ (IQR $=0.07\text{--}0.11$), while \da yields $0.21$ (IQR $=0.19\text{--}0.22$). 
\lbi significantly outperforms \da in both regimes. 
However, \da still captures the qualitative nature of the dynamics, correctly reproducing polarization and consensus patterns (see \Cref{fig:reconstruction_clean_sorted} for errors on sorted states, which highlight qualitatively good reconstructions).
This robustness is notable given that \da
\rev{has no explicit knowledge of the BCM, as it operates in a model-agnostic way}.

\paragraph{Robustness to mis-specification.}
\label{sec:robustness}
\begin{figure*}[!h]
  \centering
  \includegraphics[width=0.7\textwidth]{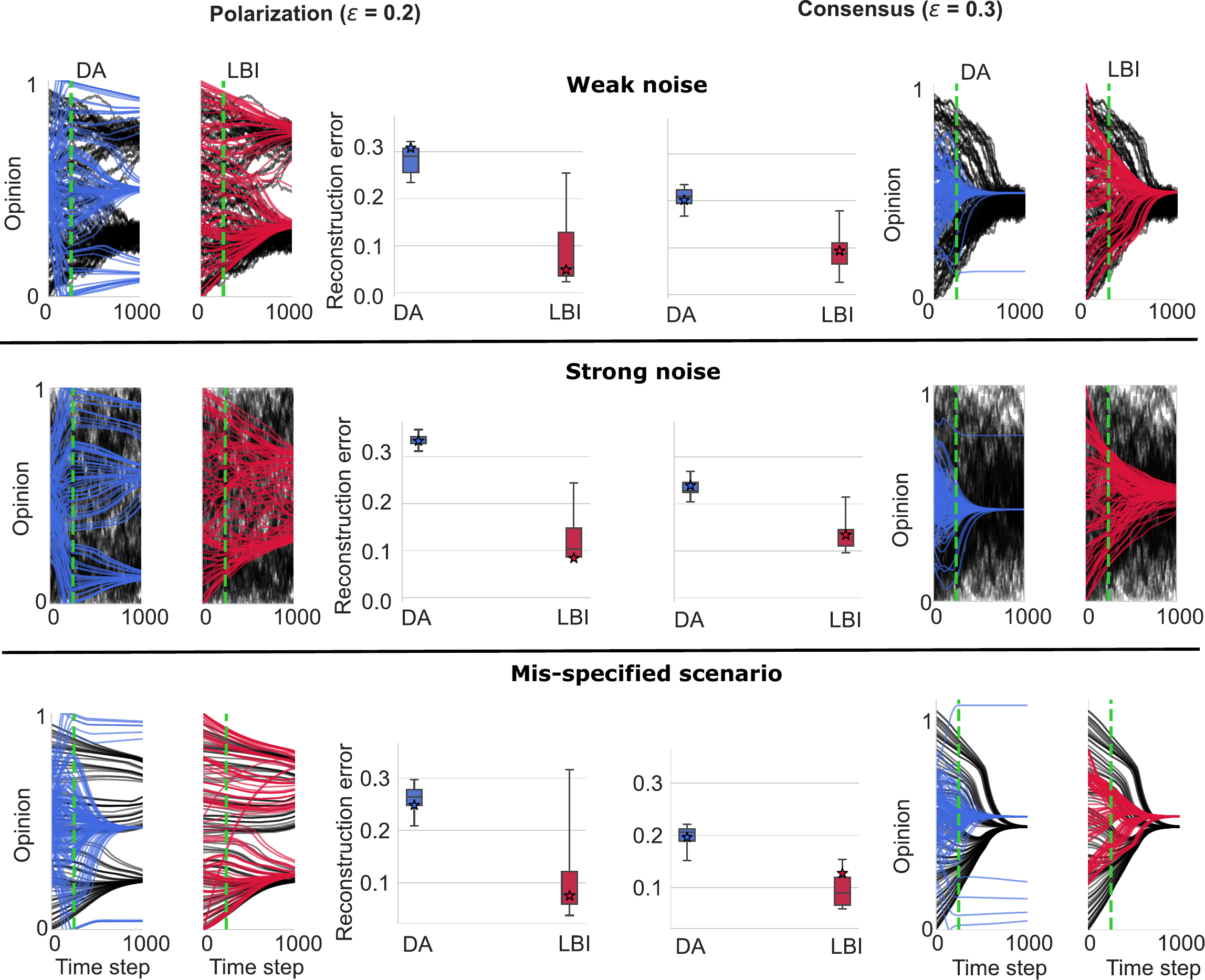}  
  \caption{Estimation results for \lbi and \da under noisy and mis-specified scenarios. 
Each row corresponds to a different setting: weak noise ($\sigma = 0.0004$, top), strong noise ($\sigma = 0.0016$, middle), and mis-specified confidence bound $\cb$ (bottom, polarization and consensus regimes swapped).
Otherwise, settings are identical to those of \Cref{fig:reconstruction_clean}.}
  \label{fig:reconstruction_misspecified}
\end{figure*}

In \Cref{fig:reconstruction_misspecified}, we examine robustness when the latent trajectories deviate from the idealized model, either due to noise in the agent states or parameter mis-specification.
We refer to \Cref{fig:all_traces} to observe the original noisy trajectories to be reconstructed.
With weak noise ($\sigma = 0.0004$, top row), the underlying trajectories remain visible, and both methods capture the overall dynamics, though \lbi achieves substantially lower errors. 
For polarization, the mean reconstruction error under \lbi is $0.09$ (IQR $=0.04\text{--}0.13$), compared to $0.28$ (IQR $=0.26\text{--}0.31$) for \da. 
For consensus, \lbi again improves performance with an error of $0.09$ (IQR $=0.07\text{--}0.11$), versus $0.21$ (IQR $=0.19\text{--}0.22$) for \da. 

Under strong noise ($\sigma = 0.0016$, middle row), where the true trajectories are almost entirely obscured, errors increase for both methods, but \lbi remains clearly superior. 
In the polarization regime, \lbi yields $0.13$ (IQR $=0.09\text{--}0.15$) versus $0.33$ (IQR $=0.33\text{--}0.34$) for \da, while in the consensus regime the errors are $0.14$ (IQR $=0.11\text{--}0.15$) and $0.23$ (IQR $=0.22\text{--}0.25$), respectively.
These results are expected noting how the ground truth trajectories are affected by such high level of noise (see (see \Cref{fig:all_traces}).

Finally, when the confidence bound $\cb$ is mis-specified (bottom row, polarization and consensus swapped between the true and inferred models), \lbi still achieves lower reconstruction errors and adapts the inferred dynamics to the observed interactions by effectively compressing the opinion space (see the bottom-right panel).
In the mis-specified scenario, the mean reconstruction error was $0.26$ (IQR $=0.25\text{--}0.28$) for \da and $0.11$ (IQR $=0.06\text{--}0.12$) for \lbi when using $\epsilon=0.3$ (consensus) to predict $\epsilon=0.2$ (polarization). 
In the reverse case, errors were $0.20$ (IQR $0.19\text{--}0.21$) for \da and $0.09$ (IQR $0.07\text{--}0.12$) for \lbi. 
These values are very similar to those obtained under correct specification.
Interestingly, \da reconstructions in this setting are slightly improved relative to the well-specified case, though they remain less accurate than those of \lbi. Overall, these results show that while both methods degrade under strong noise, \lbi is consistently more robust to both observational noise and structural mis-specification. 

Inspecting the individual traces from  \Cref{fig:reconstruction_clean,fig:reconstruction_misspecified}, \lbi faithfully recovers the true initial conditions, producing trajectories that remain close to the ground truth. 
In contrast, \da yields approximate reconstructions that nevertheless capture the main qualitative features of the dynamics. 
This \rev{finding suggests a} trade-off in which \lbi excels at precise state recovery, while \da provides coarser but still reliable reconstructions of the system’s macroscopic behavior.

\subsection{Forecasting agent interactions}
\label{sec:forecasting}

\begin{figure}[!ht]
    \centering
    \includegraphics[width=1\columnwidth]{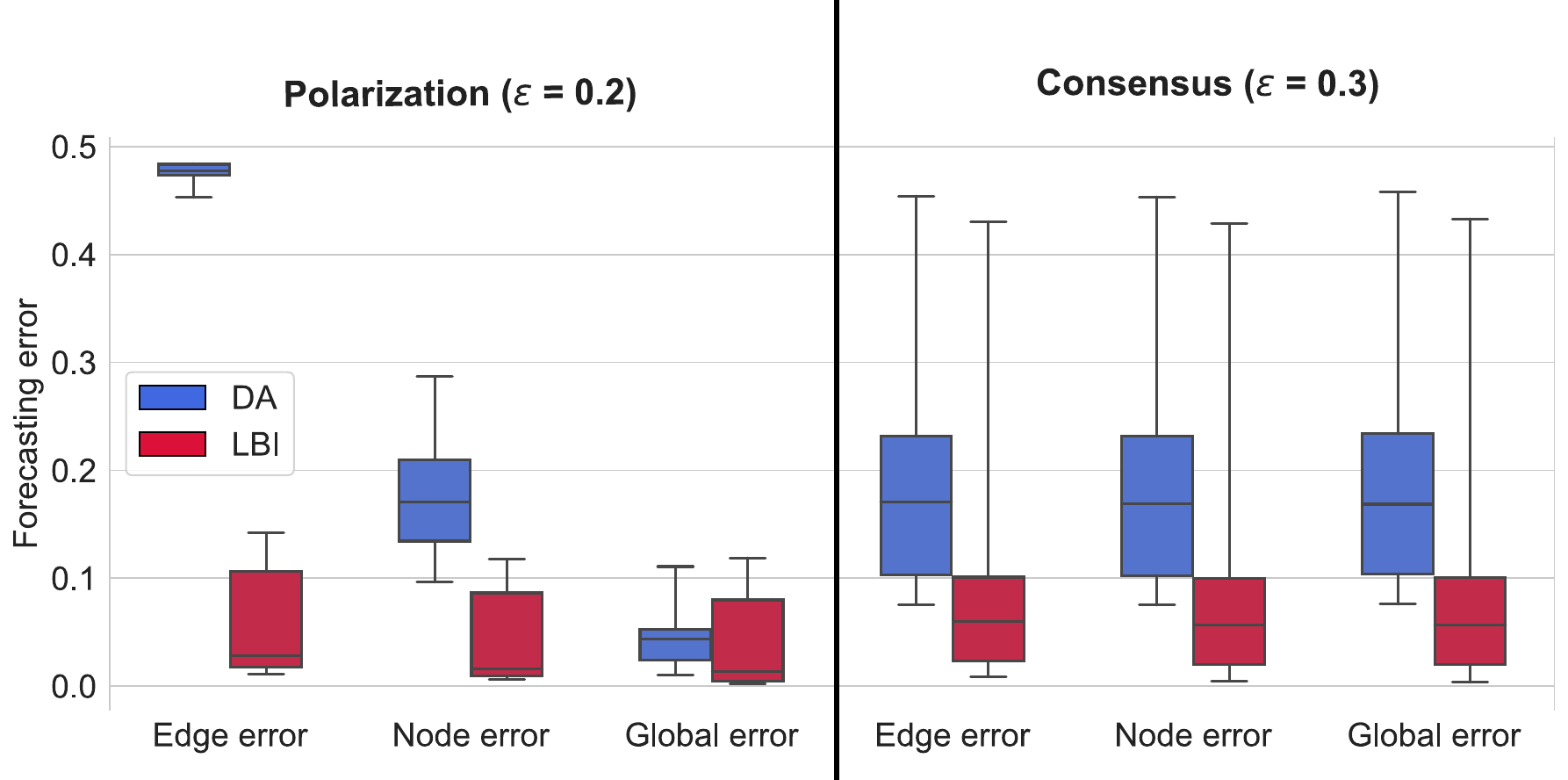} 
    \caption{Forecasting accuracy of \da and \lbi in the observation space. 
Boxplots show normalized mean absolute error (MAE) at three levels of granularity: edge (interaction-level), node (number of interactions per agent), and global (total number of interactions over time). 
Results are reported for polarization ($\cb=0.2$, left) and consensus ($\cb=0.3$, right). 
In polarization, \lbi achieves substantially lower forecasting errors across all metrics, while in consensus, the gap narrows, reflecting that a shared latent space makes reconstruction easier.}
    \label{fig:forecast_accuracy_mae_norm}
\end{figure}

Beyond reconstructing latent trajectories, we evaluate how well each method forecasts system behavior in the observation space. 
Specifically, we assess the predictive accuracy on out-of-sample intervals ($t > 250$) at three levels of granularity: interaction-level (edge error), agent-level (node error), and system-wide (global error) (see Section~\ref{sec:evaluation} for details). 
We summarize our results in \Cref{fig:forecast_accuracy_mae_norm} (see also \Cref{fig:forecast_trajectories} for full forecast trajectories of both methods and \Cref{fig:forecast_accuracy_brier} for results measured with Brier score).

As expected, \lbi achieves lower forecasting errors than \da across all metrics, particularly in the polarization scenario.
At the interaction level, \da reaches a mean error of $0.48$ (IQR $=0.47\text{--}0.48$) compared to $0.06$ (IQR $=0.02\text{--}0.11$) for \lbi. 
At the agent level, the errors are $0.17$ ($0.13\text{--}0.21$) versus $0.05$ ($0.01\text{--}0.09$), respectively. 
In the consensus regime ($\cb=0.3$), both methods improve, and the gap narrows: \da attains a mean error of $0.20$ (IQR $=0.10\text{--}0.23$), while \lbi achieves $0.12$ (IQR $=0.02\text{--}0.10$). 
These differences reflect a fundamental contrast between regimes: in consensus, all agents converge toward the same latent space, which simplifies reconstruction and forecasting. 
As a result, \da—despite lacking model-specific likelihoods—has a much easier time tracking the dynamics and becomes competitive with \lbi, particularly at aggregated (global) levels.

Importantly, \da remains robust despite \rev{being model-agnostic}. 
Without relying on the explicit likelihood, \rev{in our case study}, it still reproduces key patterns of the observed dynamics, and at coarse levels of aggregation, its forecasts often approach those of \lbi. 
This flexibility \rev{suggests \da suitability} in scenarios where the true generative model is unknown or only partially specified. 
By contrast, \rev{in our case} \lbi offers sharper precision, particularly under more complex dynamics like the polarization scenario, where the reconstruction problem is more demanding.

\FloatBarrier

\section{Discussion}
\label{sec:discussion}

This study presents the first systematic comparison of \dalong (\da) and \lbilong (\lbi) for latent state estimation in \rev{an agent-based model (ABM)}.
Using the \bcmlong (\bcm) of opinion dynamics as a testbed, we demonstrate that \rev{in this context} \lbi achieves superior accuracy in recovering agent-level latent states and generating individual-level forecasts, even under model mis-specification.
By contrast, \da remains competitive for aggregate-level predictions, accurately capturing the macroscopic dynamics of polarization or consensus despite its model-agnostic formulation.
This divergence \rev{points to} a fundamental methodological trade-off: while \lbi delivers higher precision through model-specific likelihoods, \da offers broader applicability through ensemble-based approximations that require no explicit likelihood function.
\rev{Based on our case study,} practitioners should thus weigh the need for agent-level precision against methodological complexity and model compatibility.

Interestingly, \rev{in our case} noise-corrupted latent states only degrade performance at high intensity levels.
This result suggests that both methods tolerate moderate stochasticity inherent in real-world systems.
More notably, \lbi exhibits unexpected robustness to parametric mis-specification of the confidence bound.
When the true confidence bound ($\cb = 0.3$) exceeds the assumed value ($\cb = 0.2$) during inference, \lbi infers opinions closer together than reality.
This `shrinkage' effectively compensates for the narrower assumed interaction threshold, thus ensuring agents remained within inferred \cb-neighborhoods during updates.
Consequently, while absolute state estimates diverge, the dynamics remain probabilistically consistent with observations: \lbi can implicitly correct parametric errors through state adaptation.
However, this resilience relies on symmetries inherent in the \bcm (e.g., invariance around $\state = 0.5$).
Whether similar adaptability extends to ABMs lacking such symmetries---such as those with heterogeneous agents or asymmetric interaction rules---remains an open question worth exploring.
\rev{
More generally, the broad applicability of \lbi to the plethora of heterogeneous ABMs in the literature is an active line of inquiry.
The derivation of the likelihood function is the main technical hurdle to applying \lbi~\citep{borile2025bias}.
Nevertheless, this technique has found applications to a wide range of opinion dynamics models~\citep{lenti2024likelihoodbased}, and even to complex ABMs from economics~\citep{monti2023learning}.
Moreover, when the likelihood is too hard to derive analytically, there exist alternative approximation techniques from machine learning, such as variational inference~\citep{lenti2024variational} \rev{or surrogate modeling \citep{cozzi2025learning}.}
Such techniques are easier to apply and do not require an explicit analytical deviation of the likelihood function.
}

While \da performs remarkably well for aggregate forecasting, its applicability to complex ABMs still faces important limitations.
The Ensemble Kalman Filter assumes linearizable dynamics and Gaussian uncertainties, which are often misaligned with the high-dimensional, discrete, or heterogeneous state spaces characteristic of social ABMs. \rev{Despite these theoretical mismatches, the EnKF has successfully reconstructed latent states in high-dimensional nonlinear geophysical systems \citep{houtekamer2016review, vetra2018state}, and has more recently been applied to ABMs calibrated with real-world data \citep{cocucci2022inference,suchak2024coupling, oswald2025agent}. These results suggest that the practical scope of \da may extend beyond its formal assumptions. However, its performance in social ABMs remains context-dependent, requiring systematic empirical validation and methodological adaptations tailored to discrete, heterogeneous, and strongly nonlinear settings.}

As a proof of concept, our analysis is limited to one ABM (\bcm).
While illustrative, broader generalizations require validation across diverse ABMs (e.g., in economics, epidemiology, or ecology).
\rev{In addition, it would be interesting to examine the role of data availability, which in this setting increases both the number of observations and the number of latent variables, and may therefore generate offsetting effects on inference performance.}
Nevertheless, this work fills a critical gap in the ABM calibration literature by providing the first structured comparison of \da and \lbi for latent state inference \rev{in an ABM}.
We hope it catalyzes more systematic evaluations, ultimately informing standardized calibration pipelines for computational social science.

\section{Methods}
\label{sec:methods}

\subsection{\bcmlong}
\label{sec:bcm}

The opinion dynamics model used in this work is an Agent-Based Model within the family of the \bcmlong (\bcm).
In particular, we use the deterministic, synchronous version of the model by~\citet{deffuant2000mixing}.
The Deffuant model, as all {\bcm}s, captures the phenomenon whereby individuals tend to be influenced only by those whose opinions are close to their own, reflecting a ``bounded confidence'' assumption.
This model operates according to the following key principles:
\begin{squishlist}
\item Agents and opinions: Each agent holds an opinion, represented as a real number within a specified range.
Without loss of generality, the range is typically $[0, 1]$, where the two extremes represent the polar opposites along some axis (e.g., left-right political spectrum, pro-anti abortion or gun control, or skeptical-believer in human-made climate change).
The initial opinions are distributed uniformly at random across this range.
\item Confidence Bound Parameter (\cb): This is the confidence threshold.
An agent only considers the opinions of others within an \cb-distance of their own opinion.
This distance defines the ``social neighborhood'' of each agent.
\item Opinion updating: As most other ABMs, the model works in discrete rounds.
At each iteration, each agent updates its opinion proportionally to the difference between their opinions within its \cb-bound neighborhood.
This mechanism models the idea that agents are influenced by others within their confidence bound.
\item Convergence dynamics: Over time, the population tends to evolve toward clusters of consensus.
Multiple clusters may form if the initial opinions are too dispersed or if \cb is relatively small, thus leading to polarized groups with limited influence across clusters, or even complete opinion fragmentation.
\end{squishlist}
The Deffuant model has been instrumental in understanding how social fragmentation, echo chambers, and consensus can emerge from simple local interaction rules.
It is widely used in modeling political polarization, the spread of ideas, and behavior on social networks.

More formally, the update equations of the Deffuant models are as follows.
Let $\state(t) \in [0,1]^N$ represent the opinion vector of the system of $N$ agents at time $t$.
Let $\statei{i}(t)$ represent its $i$-th component, i.e., the opinion of agent $i$ at the same time.
Let the bounded confidence threshold be $0 < \cb \leq 1 $.
Agent $i$ considers the opinions of agent $j$ only if $\lvert \statei{i}(t) - \statei{j}(t) \rvert \leq \cb$.
We denote this condition with an indicator function $\obsi{ij}(t) = \mathbb{1}\left(|\statei{i}(t) - \statei{j}(t)| \leq \cb\right)$, which we call the \emph{observation}.
At each time step, the opinion of agent $i$ is updated based on the opinions of all agents within its bounded confidence interval.
Let $\neighbors_i(t)$ denote the set of agents whose opinions fall within agent $i$'s confidence interval at time $t$ (\emph{confidence set})
\[
\neighbors_i(t) = \{ j : \left\lvert \statei{i}(t) - \statei{j}(t) \right\rvert \leq \cb \}.
\]
Then, the opinion update rule for each agent $i$ is
\begin{align}
\label{eq:update}
\statei{i}(t+1) = \statei{i}(t) + \rate \sum_{j \in \neighbors_i(t)} \left( \statei{j}(t) - \statei{i}(t) \right),
\end{align}
where $\rate\in [0,0.5]$ is a \emph{convergence rate} parameter that regulates the speed of the convergence of the system and its dynamics, but does not affect its steady state as this version of the model is synchronous~\citep{lanchier2012critical,li2024straightforward} (differently from the original asynchronous model where a single pair of agents interacts at each time step~\citep{deffuant2000mixing}).

Note that the confidence set depends on the absolute value of the difference between agents' opinions, and thus is invariant to translation (modulo border effects) and reflection.
Similarly, the opinion updates are proportional to the signed difference in opinions.
Thus, the opinion trajectories of a system with initial opinion vector \state and the ones with initial opinion vector $\state' = 1-\state$ are symmetric around the midpoint of the opinion space $\statei{}=0.5$.
As such, they are indistinguishable when we only have access to the observations $\obs$, and we are unable to directly observe the opinions of the agents \state, as explained next.

\subsection{Latent and Observable Variables}
\label{sec:lat-obs}

Data-driven agent-based models (ABMs) are attractive due to their flexibility in simulating complex systems of interacting agents while achieving strong performance in real-world applications \citep{wiese2024forecasting}.
However, we rarely observe agent-level states directly, as they often represent hard-to-measure constructs.
Instead, we might observe agent-level actions or summary statistics of the system and track them over time.

For example, consider measuring user interactions over time on a social platform such as Reddit.
This system represents an evolving interaction network---who replies to whom and when---while the underlying opinions or attitudes driving these interactions are latent~\citep{monti2020learning}.
A model such as the \bcm captures the evolution of these latent opinions that are not directly accessible from the data.
This naturally leads us to distinguish between \emph{observable} variables, denoted~by~$\obs$---such as the evolving interaction network---and \emph{latent} variables, denoted~by~$\state$---such as the agent's opinions described by the \bcm.

Formally, we can think of our ABM as a dynamical system $\mathbf{f}$ that describes the evolution of the (latent) agent states
\begin{equation}
    \state(t) = \mathbf{f}\left(\state(\tau \leq t)\right),
    \label{eq:ds}
\end{equation}
where $\state(t)$ represents the agent states at times $t$ and $\state(\tau \leq t)$ the full history up to that point.

Observations are then derived through an observation operator, $\mathbf{h}$, applied to the latent states at a given time
\begin{equation}
    \obs(t) = \mathbf{h}( \state(t) ).
    \label{eq:obs_op}
\end{equation}
In general ABMs, $\state(t)$ might also depend on $\obs(\tau < t)$.
Here, we assume that
that \obs is a measure of \state, so it does not affect its dynamics.

In our previous \bcm example, the opinions of the agents represent the latent state $\state$, the network interactions are the observable variables $\obs$, the update \Cref{eq:update} describes the system $\mathbf{f}$, and the confidence bound \cb determines the observation operator $\mathbf{h}$.
Our main goal is to infer the latent trajectory $\state = (\state(0), \ldots, \state(t))$ from the observed data $\obs = (\obs(0), \ldots, \obs(t))$.
We can frame this task as computing the posterior distribution $p(\state \mid \obs)$ according to Bayes rule
\begin{equation}
    p(\state \mid \obs) \propto p(\obs \mid \state) p(\state),
    \label{eq:bayes}
\end{equation}
where $p(\obs \mid \state)$ is the \emph{likelihood} of the observed data given the model and $p(\state)$ is the \emph{prior} over the ABM trajectories.
If the ABM is deterministic, the prior reduces to the distribution over initial states $\state(0)$.
The posterior provides a unifying theoretical foundation from which different inference strategies have been developed~\citep{kolic2022estimating,monti2020learning,lenti2024variational}.
This work focuses on two main approaches: \dalong (\da) and \lbilong (\lbi).

\da focuses on approximating the full posterior distribution by making assumptions about the likelihood and the model structure.
This approach relaxes the need to craft a model-specific likelihood.
However, the quality of \da depends on how well the approximated likelihood and structure match the true data-generating process.
\Cref{sec:da} discusses the specifics of \da.

In contrast, \lbi seeks the \emph{most likely latent state trajectory} \xest by maximizing the log-likelihood
\begin{equation}
    \label{eq:lbi-general}
    \xest = \argmax_{ \state } \log p( \obs \mid \state ).
\end{equation}
The main challenge here is deriving a model-specific likelihood~\citep{lenti2024likelihoodbased}.
When successful, the maximum likelihood trajectory will represent a realistic explanation of the data stemming from the latent states.
Note that the maximum likelihood estimate can be understood in terms of Bayesian statistics as the maximum a posteriori (MAP) estimate, i.e., the mode of the posterior distribution, when the priors are non-informative.
\Cref{sec:lbi} discusses the details of \lbi.

\subsection{\dalong: The Ensemble Kalman Filter}
\label{sec:da}

The Ensemble Kalman Filter (EnKF) is a widely used data assimilation (\da) technique for estimating the latent states of dynamical systems given noisy observations~\citep{houtekamer2016review}. \rev{In general, \da integrates real-time observations with model simulations to create an optimized estimate of the true state of the system.}
In the context of the \bcm, these methods provide a practical approach to infer individual agent opinions $\state(t)$ based on observed interactions $\obs(t)$.
In contrast to likelihood-based methods that require explicit computation of the likelihood function, the EnKF employs an ensemble of model realizations to approximate likelihood and state distributions and update estimates recursively.

The EnKF is a sequential filtering approach that estimates the latent state of a system by propagating an ensemble of initial conditions sampled from a prior distribution through the system dynamics and updating the ensemble members' estimates based on new observations.
It consists of two main steps: the \emph{forecasting step}, \rev{where the system is driven forward in time through model simulation,} and the \emph{analysis step}, \rev{where the forward-driven state is corrected based on how much it deviates from the observation}. \rev{To differentiate between these steps, we will use diacritic $\xestf$ to refer to states in the forecasting step, and hatted $\xest$ for the state estimate in the analysis step.}

First, we sample an ensemble of \rev{$N_e$} initial states $\hat{\state}^k(t) \sim p_\state$, where $p_\state$ is some prior distribution, and $\xest^k(t)$ represents the estimated state for ensemble member \rev{$k$} at time $t$.

In the forecasting step, each ensemble member propagates independently according to the \bcm model:
\begin{equation}
     \xestf^k(t) := \mathbf{f}( \xest^k(t-1) ) + \eta_i, \quad \rev{k} = 1, \dots, N_e,
     \label{eq:enkf_forecast}
\end{equation}
where $\mathbf{f}(\state)$ represents the \bcm's opinion update (\Cref{eq:update}), $\eta_i \sim \mathcal{N}(0, \mathbf{\Sigma}(t))$ is the model uncertainty represented by unbiased Gaussian noise with covariance matrix $\mathbf{\Sigma}(t)$.%

We perform the analysis step when the observation $\obs(t)$ becomes available by updating the ensemble members according to the correction matrix, $\mathbf{K}(t)$, called the \emph{Kalman gain}:
\begin{equation}
    \xest^k(t) = \xestf^k(t) + \mathbf{K}(t) \left[ \obs(t) - \mathbf{h}( \xestf^k(t) ) \right] + \nu_i,
    \label{eq:enkf_analysis}
\end{equation}
where $\mathbf{h}$, the observation operator from \Cref{eq:obs_op}, maps the latent opinions to observed interactions, and $\nu_i \sim \mathcal{N}(0, \mathbf{R}(t))$ is a Gaussian perturbation term with the covariance matrix $\mathbf{R}(t)$ representing the observational noise.
\rev{We can write the \textit{Kalman gain} in matrix form as} 
\rev{\begin{equation}
    \mathbf{K}(t) =  \mathbf{A}(t) \mathbf{S}(t)^T \left[ \mathbf{S}(t) \mathbf{S}(t)^T + \mathbf{R}(t) \right]^{-1},
    \label{eq:enkf_gain}
\end{equation}}
\rev{where $\mathbf{A}(t) = [ \xestf^k(t) - \langle \xestf^k(t) \rangle_k ]$ is the matrix of zero-mean ensemble members, and $\mathbf{S}(t) = [ h(\xestf^k(t)) - \langle h(\xestf^k(t)) \rangle_k ]$ is the zero-mean observation matrix.}
The estimated latent state at time $t$ is the average over the ensemble:
\begin{equation}
    \xest(t) = \langle \xest^k(t) \rangle_k,
    \label{eq:da_est}
\end{equation}
and the approximated posterior of the trajectory is given by the higher moments over the ensemble of particles.

Given the new observation $\obs(t)$, the Kalman gain determines how much we should adjust the forecasted state $\xestf(t)$ to obtain the corrected estimate $\xest(t)$.
In Bayesian terms, this corresponds to an optimal trade-off between the prior uncertainty (from the forecast step) and the observational uncertainty.
The update step effectively applies Bayes' rule in a Gaussian setting, yielding the new state estimate by weighting the prior state estimate and the likelihood of the observation.
This results in a posterior distribution that balances prior knowledge with new data.

This approach allows EnKF to efficiently incorporate new observations while accounting for model uncertainty and ensemble spread.
However, it relies on the assumption of Gaussian uncertainties and \rev{sequentially} linearizable dynamics to operate optimally. \rev{Importantly, only local linearizability is required at the observation times. Therefore, the EnKF can successfully handle strongly nonlinear dynamics, provided observations are frequent enough to prevent substantial divergence between successive updates \citep{vetra2018state}.}
In the \bcm setting, the observation operator~$\mathbf{h}$ matching agent opinions to interactions is highly non-linear, as interactions occur according to a step function of the distance of agents' opinions.
Despite this limitation, we use the EnKF to estimate opinions and evaluate its performance compared to \lbi.

\subsection{Likelihood-Based Inference: Gradient Descent with Automatic Differentiation}
\label{sec:lbi}

The problem of parameter estimation in opinion dynamics can be approached as a \emph{likelihood-based inference} task, where the goal is to infer the initial opinions of agents from observed interactions over time.
The \bcmlong by Deffuant describes the evolution of opinions based on pairwise interactions, governed by a confidence bound \cb and a convergence rate \rate.
Rather than simulation---that can be seen as the \emph{forward} pass---we estimate $x(0)$ given observed interaction data.
Since opinions are latent and only interactions are observed, the solution involves optimizing $x(0)$ such that the resulting opinion trajectories best explain the observed interaction patterns \obs, as expressed by \Cref{eq:lbi-general}.

We formulate the task as a \emph{maximum likelihood estimation (MLE)} problem, where the probability of observed interactions is modeled as a function of pairwise opinion differences.
We derive a differentiable loss function by using the log-likelihood of the interaction data, which can be seen as a \emph{binary cross-entropy loss} comparing predicted interaction probabilities to observed interactions.
In fact, the binary cross-entropy corresponds exactly to the log-likelihood of the system under the estimate $\xest$:

\begin{align}
\begin{split}
    \log p( \obs \mid \state ) &=
    \sum_{i, j} \Big(
    \obs_{i, j}
        (\left\lvert \xest_{i}(t) - \xest_{j}(t) \right\rvert - \cb) \\
    &+ (1 - \obs_{i, j})
        (1 - \left\lvert \xest_{i}(t) - \xest_{j}(t) \right\rvert - \cb)
    \Big)
\end{split}
\end{align}

This setting makes the problem similar to a supervised learning task, where the estimated logits $(\left\lvert \xest_{i}(t) - \xest_{j}(t) \right\rvert - \cb)$ can be seen as a score used to predict whether the agents interact.

These logits, once we fix the parameters of the model, are purely a function of \xest, which in turn is a deterministic function of the past interactions and of $\xest(0)$.
The latent states $\state(0)$, in fact, define the entire opinion trajectory via the \bcm update equations.
At this point, the whole training consists in optimizing $\xest(0)$ via \emph{gradient-based methods} such as \emph{RMSprop} or \emph{Adam}.

Auto-differentiation enables efficient computation of gradients through the sequence of opinion updates, ensuring stable and effective convergence.
However, in order for this solution to work efficiently, it requires to rewrite the update of opinions as a tensorial operation.
In other words, 
we represent the update process as a sequence of vectorized operations rather than iterating over individual agents.
Given the initial opinion vector $\state(0)$, the opinion evolution can be expressed as a sequence of transformations governed by a time-dependent adjacency structure.
At each time step $t$, this interaction structure is represented as a sparse $N \times N$ adjacency matrix $A(t)$, s.t. $A_{i,j}(t) = 1$ iff $j \in \neighbors_i(t)$.
This allows us to rewrite the opinion update rule from \Cref{eq:update} as the matrix operation
\[
\state(t+1) = \rate A(t)^\top \state(t) + 
    (\mathbf{1} - \rate A(t)^\top \mathbf{1}) \circ \state(t),
\]
where $\mathbf{1}$ is a vector of ones, and $\circ$ represents the Hadamard (element-wise) product.
This formulation highlights the two key components of the update: (i) each agent retains a fraction of its previous opinion, weighted by the total influence received, and (ii) each agent absorbs a weighted sum of the opinions of its influencing neighbors.
By iterating this transformation over $T$ steps, the full opinion trajectory can be expressed as a deterministic function of the initial condition.
This vectorized formulation enables efficient computation using tensor operations, thus making it well-suited to auto-differentiation and optimization within machine-learning frameworks such as PyTorch.
This way, the loss function computed from an interaction at time $t$ will be back-propagated over each opinion vector, defined as a function of the previous one, and finally updating the free variable $\state(0)$.
Finally, to stabilize results and avoid convergence to suboptimal local minima, multiple random restarts can be employed during optimization, as well as standard gradient-descent regularization techniques such as weight decay.

\subsection{Evaluation Protocol}
\label{sec:evaluation}

We run the model described in \Cref{sec:bcm} for various settings, to generate a variety of ground-truth data traces that we use to evaluate the performance of \da and \lbi.
The model has two parameters: the confidence threshold \cb and the convergence rate \rate.
The latter simply determines the speed of convergence of the system to the steady state, so we choose $\rate=10^{-4}$ in order to have sufficiently slow convergence to be able to observe the transient and learn the agent states both with \da and \lbi.
The confidence threshold \cb is instead fundamental to the ABM dynamics.
We choose two values of \cb that lead to different behaviors.
Specifically, we run simulations with $\cb=0.2$, which is small enough to generate multiple opinion clusters (\emph{polarization scenario}), and we also consider a larger $\cb=0.3$, which is large enough to make agents converge on a single consensus opinion (\emph{consensus scenario}).
In addition to these two parameters, we explore the effect of adding a noise term to the opinion update, \Cref{eq:update}.
In addition to the deterministic scenario described in \Cref{sec:bcm}, we consider five noise levels, ranging from $2^0 \times 10^{-4}$ to $2^4 \times 10^{-4}$.
These magnitudes encompass all noise magnitudes from a small alteration of the deterministic trajectories to completely noisy dynamics (these values should be interpreted in comparison with the convergence rate $\rate=10^{-4}$).
Finally, to explore the effect of stochasticity in the model, we consider 10 seeds of the random number generator, which govern both the initial conditions of the opinions $\state(0)$ and the realizations of the noise.
Summarizing, we run the ABM for 2 values of \cb, 6 values of noise (including zero noise), and 10 different random seeds, resulting in a total of 120 distinct simulations.

We run all simulations for $N=100$ agents and a complete interaction network, in the sense that all agents potentially interact with all other agents, leading to a total of $E=N(N-1)/2=4950$ edges.
We run the ABM for a total of $T=1000$ time steps and record:
\begin{squishlist}
    \item the ground truth opinions $\state(t)$;
    \item the edge indicator variable $\obsi{ij}(t)$ for whether agents $(i,j)$ interact at time $t$;
    \item how many interactions agent $i$ has with all other agents at time t, namely node-level interactions $\obsi{i}(t)=\sum_j \obsi{ij}(t)$
    \item how many interactions all agents have with all other agents at time $t$, namely global-level interactions $y(t) = {1 \over 2} \sum_i y_i(t)$.
\end{squishlist} 

To infer the latent opinions, we use ground truth data up to time step $250$ (or $25\%$ of the simulation length) as input for both inference algorithms, \da and \lbi.
Our results do not depend strongly on the specific choice of time step $250$.
We select this time step to strike a balance between giving enough data to the inference algorithms and not having already reached convergence to the steady state.
Any other step that is neither too early or too late in the simulation would lead to similar results.
We denote estimates by $\xest(t)$.
With our latest estimates at $t=250$ for each algorithm, we then run the ABM up to the final time step $T=1000$, without any adjustment to the latent opinions, and use these data traces to evaluate the out-of-sample forecasting capabilities of both the \da and \lbi algorithms.
In the next sections, we discuss in more detail the evaluation for the inference and forecasting tasks.

To explore the effect of mis-specification, we run both the \da and \lbi algorithms with the correct value of \cb and with the mis-specified value of \cb (i.e., $\cb=0.2$ in the inference when $\cb=0.3$ in the ground truth, and vice versa).
Noise represents a further source of mis-specification: we never account for it in the inference algorithms; therefore, the larger the noise, the stronger the mis-specification.

A final dimension for evaluation is what exact ground truth data are fed to the \da and \lbi inference algorithms.
By construction, \lbi can only handle the finest-grained, edge-level observations $\obsi{ij}(t)$.
Instead, \da can use edge-level information, but also more aggregated node level ($\obsi{i}(t)$) and global level ($\obsi{}(t)$) information.
In conclusion, we infer the latent variables for 240 data traces via \lbi (120 traces with well-specified \cb and 120 traces with mis-specified \cb) and 720 data traces via \da (240 traces with edge-level, node-level, and global-level information each).

\subsubsection{Inference}

We focus on the latent opinion inference at the initial step ($t=0$) and at the final step of the inference part ($t=250$).
As an evaluation metric, we use the Mean Absolute Error (MAE).
So, the inference error at time $t$ averaged over all agents is 
\begin{equation}
    \error(t) = {1 \over N} \sum_i \left| \statei{i}(t) - \xesti{i}(t) \right|.
\end{equation}

According to the metric above, the inference algorithms must infer the correct opinion of all agents $i$.
However, as discussed in \Cref{sec:bcm}, there is a symmetry around $x=0.5$ which prevents identification of the exact value of $\state$.
This is also known as label switching symmetry in probabilistic models~\citep{stephens2000dealing}.
To account for this symmetry, we consider a symmetric version of $\error(t)$, giving 
\begin{equation}
    \error_{\text{symm}}(t) = {1 \over N} \sum_i \min \left( \left\lvert \statei{i}(t) - \xesti{i}(t) \right\rvert, \left\lvert \statei{i}(t) - \left(1-\xesti{i}(t)\right) \right\rvert  \right).
\end{equation}
In the absence of edge-level information, it may be possible to get the correct distribution of opinions but not to assign the correct opinion to the correct agent.
To compare the distributions, we sort both the ground truth and the estimate:
\begin{equation}
    \error_{\text{sort}}(t) = {1 \over N} \sum_i \left\lvert \text{sort}(\state(t))_i - \text{sort}(\xest(t))_i \right\rvert.
\end{equation}

\subsubsection{Forecasting}
To recap, using the \da- and \lbi-inferred opinions at time step $250$, we run the ABM for $750$ further time steps.
Then, we compare the obtained trace to the ground truth time series to evaluate the out-of-sample forecasting capabilities of \da and \lbi.
This measure is likely correlated to how well the methods reconstruct the latent opinions at time step $250$.
However, due to the nonlinearity of the complex system, different errors in the reconstructed opinions may lead to more severe divergence from the ground truth over the simulation.

As indicators of forecasting quality, we may be interested in how well the inference algorithms reconstruct the edge-level interactions $\obsi{ij}(t)$, the node-level interactions $\obsi{i}(t)$, or the global interactions $\obsi{}(t)$.
When considering edge-level interactions, we quantify the forecasting error by the MAE.
Therefore, letting $\widehat{p}_{ij}(t)$ denote the forecast probability of an interaction between $(i,j)$ at $t$ when starting the ABM from the inferred opinions at time step $250$, the MAE score at time $t$ is
\begin{equation}
    \mathcal F_{\text{edge}}(t) = {1 \over E} \sum_{ij} \left\lvert \obsi{ij}(t) - \widehat{p}_{ij}(t) \right\rvert.
\end{equation}
We choose the MAE as the main evaluation metric, which is an axiomatically good scoring metric for quantification tasks~\citep{sebastiani2020evaluation}.

To evaluate how well we match node-level interactions, we aggregate the probabilities of interactions at the node level both in the ground truth and in the simulations following from latent opinions inference, leading to a probabilistic count error metric
\begin{equation}
    \mathcal F_{\text{node}}(t) = {1 \over N} \sum_{i} \left\lvert \obsi{i}(t) - \sum_j \widehat{p}_{ij}(t) \right\rvert.
\end{equation}
Finally, the match to total interactions is computed by summing all probabilities of interactions, giving
\begin{equation}
    \mathcal F_{\text{global}}(t) =  \left\lvert  \obsi{}(t) - \sum_{ij} \widehat{p}_{ij}(t) \right\rvert.
\end{equation}

\paragraph{Data availability.} The study is based on synthetic data. All datasets generated and analyzed during this study, along with the analysis code, will be made available at https://doi.org/10.5281/zenodo.17153273.

\bibliographystyle{abbrvnat}
\bibliography{biblio}

\clearpage

\appendix
\section{Appendix}
\setcounter{figure}{0}
\renewcommand{\thefigure}{A\arabic{figure}}

In this appendix, we present additional figures that complement the results discussed in the main text.

\subsection{Latent State Reconstruction}

\begin{figure}[htb]
    \centering
    \begin{minipage}{0.23\textwidth}
        \centering
        \includegraphics[width=\linewidth,clip,trim=0 240 0 0]{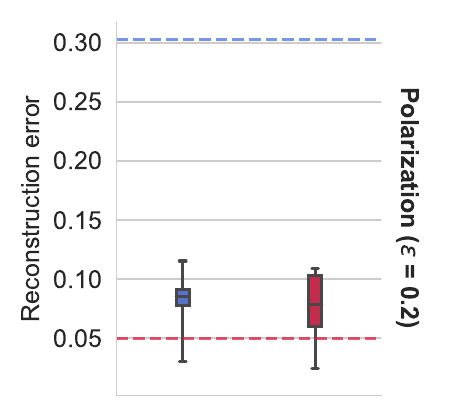}
        \makebox[\linewidth]{\hspace{1em} \scriptsize\sffamily DA \hspace{2em} LBI}
    \end{minipage} %
    \begin{minipage}{0.23\textwidth}
        \centering
        \includegraphics[width=\linewidth,clip,trim=0 25 0 215]{figure_supp3}
        \makebox[\linewidth]{\hspace{1em} \scriptsize\sffamily DA \hspace{2em} LBI}
    \end{minipage}
    \caption{Reconstruction error using the sorted metric, which compares opinion distributions without enforcing identity of agents, for the polarization (left) and consensus (right) scenarios without any mis-specification. Dashed horizontal lines indicate the median error measured by the unsorted error metric (shown in \Cref{fig:reconstruction_clean}).}
    \label{fig:reconstruction_clean_sorted}
\end{figure}

\paragraph{Aggregate reconstruction.}
\Cref{fig:reconstruction_clean} shows results for the task of latent state reconstruction, assuming one is interested in reconstructing the latent state of each agent individually.
In other words, the true state of an agent $i$ is compared to the reconstructed state of the same agent.
However, in some contexts, one might be interested in reconstructing a faithful \emph{distribution} of latent states, disregarding their individual identity.
To measure the quality of \lbi and \da in this context, we adopt in this section a \emph{sorted metric}, where each agent's reconstructed latent state is compared against the agent with the same rank in the original set of latent states.
That is, we measure the reconstruction error between the \emph{sorted} vector of true and reconstructed latent states.
\Cref{fig:reconstruction_clean_sorted} shows these results.
In this context, the two methods perform similarly, with outcomes essentially tied in the polarization scenario and \lbi showing a slight advantage in the consensus scenario.

\begin{figure}[htb]
    \centering
    \includegraphics[width=\linewidth,clip,trim=0 35 0 0]{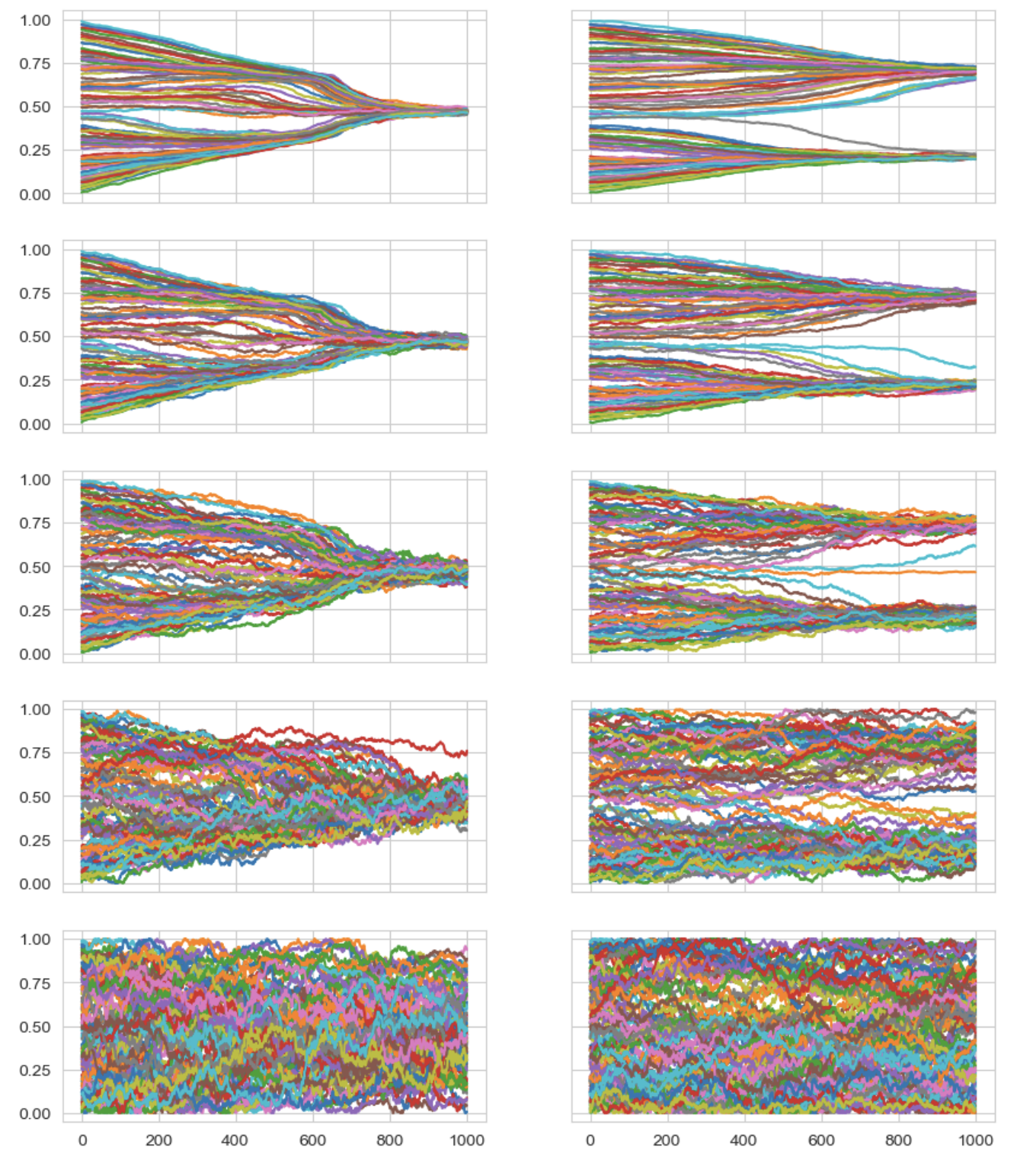}
    \makebox[\linewidth]{\hspace{1em} \tiny\sffamily Time step \hspace{14em} Time step \hspace{1em}}
    \caption{Ground-truth trajectories of the Bounded-Confidence Model under different noise intensities (top corresponds to \emph{no noise}, bottom to the \emph{strong noise} setting). Left column represents the consensus scenario and right column the polarization scenario. In each plot, we represent opinion evolution.}
    \label{fig:all_traces}
\end{figure}

\paragraph{System trajectories in the presence of noise.}
In \Cref{fig:reconstruction_misspecified}, we have shown results under a particular type of mis-specification, i.e., noise in the original latent trajectories.
To further illustrate this scenario and aid comprehension, we report in \Cref{fig:all_traces} such ground truth trajectories under our three settings of noise.

\begin{figure}[htb]
    \centering
    \begin{minipage}{0.23\textwidth}
        \centering
        \includegraphics[width=\linewidth,clip,trim=0 255 0 0]{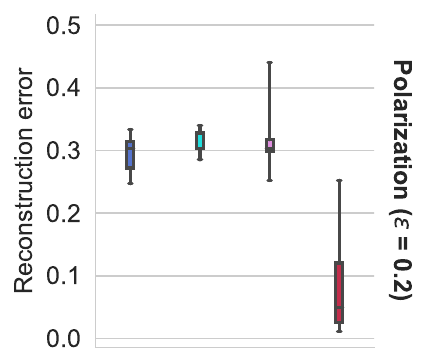}
\makebox[\linewidth]{ \hspace{1.1em}%
        \scriptsize\sffamily
  \rotatebox[origin=c]{90}{\makebox[5em][r]{DA (edge)}} \hspace{0.5em}
  \rotatebox[origin=c]{90}{\makebox[5em][r]{DA (node)}} \hspace{0.5em}
  \rotatebox[origin=c]{90}{\makebox[5em][r]{DA (global)}} \hspace{0.5em}
  \rotatebox[origin=c]{90}{\makebox[5em][r]{LBI}}%
    }
    \end{minipage} %
    \begin{minipage}{0.23\textwidth}
        \centering
        \includegraphics[width=\linewidth,clip,trim=0 60 0 200]{figure_supp2}
\makebox[\linewidth]{ \hspace{1.1em}%
        \scriptsize\sffamily
  \rotatebox[origin=c]{90}{\makebox[5em][r]{DA (edge)}} \hspace{0.5em}
  \rotatebox[origin=c]{90}{\makebox[5em][r]{DA (node)}} \hspace{0.5em}
  \rotatebox[origin=c]{90}{\makebox[5em][r]{DA (global)}} \hspace{0.5em}
  \rotatebox[origin=c]{90}{\makebox[5em][r]{LBI}}%
    }
    \end{minipage}
    \caption{Comparison of reconstruction performance when DA uses different observation operators (edge-, node-, and global-level), for the polarization (left) and consensus (right) scenarios without any mis-specification. LBI is also reported as a reference.}
    \label{fig:reconstruction_clean_multiple_da_methods}
\end{figure}

\paragraph{DA observation operators.}
In the main text, for \lbi we focused on the ``edge'' observation operator.
\Cref{fig:reconstruction_clean_multiple_da_methods} shows the performance obtained in this task by \da by each of its three possible observation operators (edge-, node-, and global-level), with \lbi also reported as reference.

\subsection{Forecasting Accuracy}
\begin{figure*}[htb]
    \centering
    \includegraphics[width=0.75\linewidth]{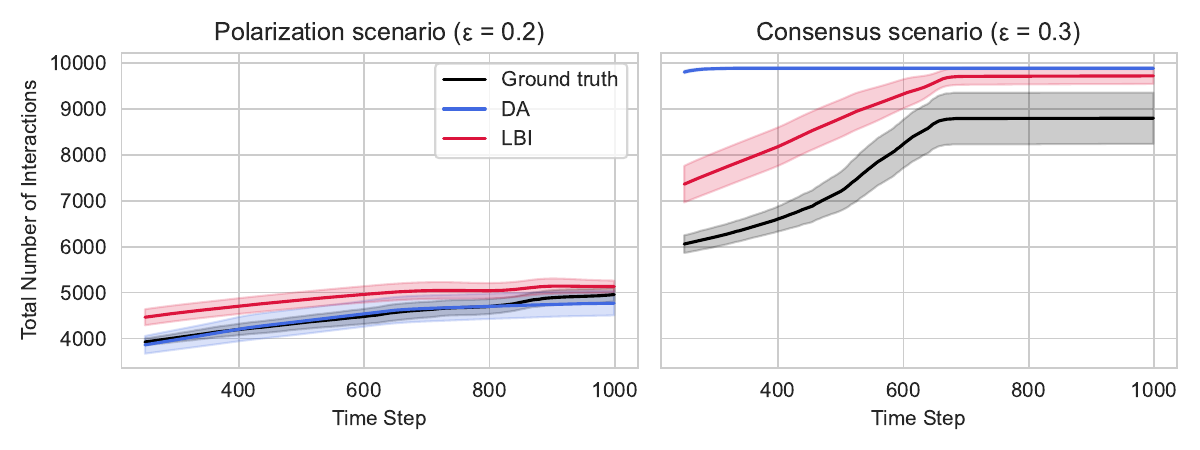}
    \caption{Forecasted system-wide trajectories of total number of interactions under (left) polarization ($\epsilon = 0.2$) and (right) consensus ($\epsilon = 0.3$) scenarios. We compare ground truth (black), DA (blue), and LBI (red). Shaded regions indicate variability across runs. In the polarization scenario, both methods qualitatively track system behavior. Instead, in the consensus scenario, only \lbi is able to make substantially accurate forecasts.}
    \label{fig:forecast_trajectories}
\end{figure*}

\begin{figure}[htb]
    \centering
    \includegraphics[width=\linewidth]{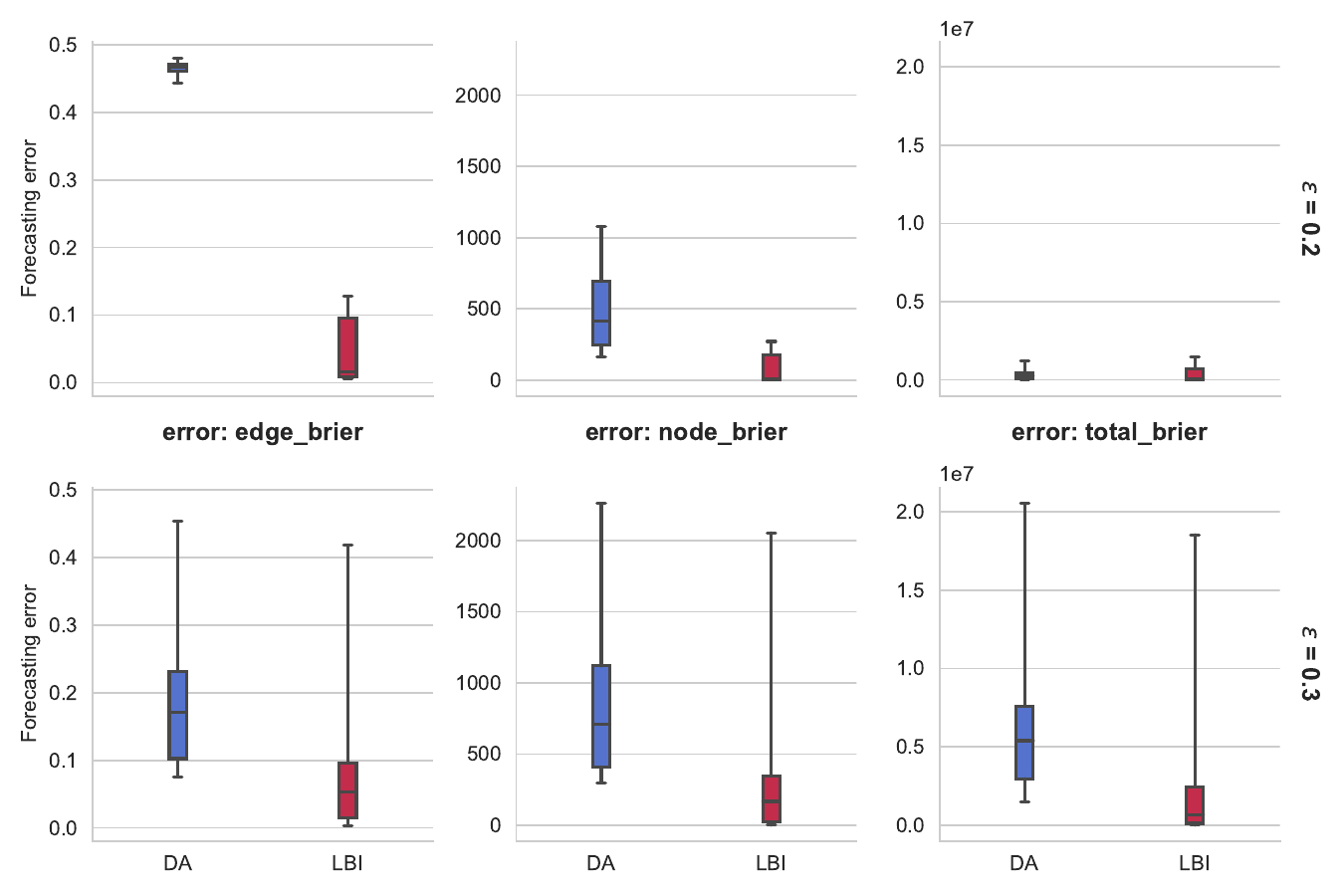}
    \caption{Forecasting accuracy measured by Brier score at three levels of granularity: edge-level, node-level, and global-level. Higher values indicate more accurate forecasts. }
    \label{fig:forecast_accuracy_brier}
\end{figure}

We next examine how well DA and LBI recover future dynamics once trained on partial trajectories. 
Here, we provide extended analyses of this forecasting task.
In \Cref{fig:forecast_trajectories}, we show explicitly the time series of the quantity to forecast (i.e., the total number of interactions) in the ground truth and in the reconstruction by \lbi and \da.
Finally, we also report additional metrics, obtained by aggregating errors at the node-, edge-, and global-level.
\Cref{fig:forecast_accuracy_brier} shows the same result of \Cref{fig:forecast_accuracy_mae_norm} but measured with the Brier score.

\end{document}